\documentclass[letterpaper, 10 pt, conference]{ieeeconf}  

\usepackage{cuted}
\usepackage{flushend}
\usepackage{cite}
\usepackage{amsmath,amssymb,amsfonts}
\usepackage{algorithm}
\usepackage{algorithmicx}
\usepackage{algpseudocode}
\usepackage{graphicx}
\usepackage{textcomp}
\usepackage{subcaption}
\usepackage{xcolor}
\usepackage{float}
\usepackage{authblk}
\newtheorem{remark}{Remark}

\def\BibTeX{{\rm B\kern-.05em{\sc i\kern-.025em b}\kern-.08em
    T\kern-.1667em\lower.7ex\hbox{E}\kern-.125emX}}
\IEEEoverridecommandlockouts                              

\title{\LARGE \bf
Incorporating System-level Safety Requirements in Perception Models via Reinforcement Learning
}
\author[1]{Weisi Fan}
\author[2]{Jesse Lane}
\author[2]{Qisai Liu}
\author[2]{Soumik Sarkar}
\author[1]{Tichakorn Wongpiromsarn}
\affil[1]{Department of Computer Science, Iowa State University}
\affil[2]{Department of Mechanical Engineering, Iowa State University}

\begin{document}

\newtheorem{lemma}{Lemma}
\newtheorem{corollary}{Corollary}
\newtheorem{Definition}{Definition} 
\newtheorem{Remark}{Remark} 

\maketitle
\thispagestyle{empty}
\pagestyle{empty}

\begin{abstract}

    Perception components in autonomous systems are often developed and optimized independently of downstream decision-making and control components, relying on established performance metrics like accuracy, precision, and recall. Traditional loss functions, such as cross-entropy loss and negative log-likelihood, focus on reducing misclassification errors but fail to consider their impact on system-level safety, overlooking the varying severities of system-level failures caused by these errors. To address this limitation, we propose a novel training paradigm that augments the perception component with an understanding of system-level safety objectives. Central to our approach is the translation of system-level safety requirements, formally specified using the rulebook formalism, into safety scores. These scores are then incorporated into the reward function of a reinforcement learning framework for fine-tuning perception models with system-level safety objectives. Simulation results demonstrate that models trained with this approach outperform baseline perception models in terms of system-level safety.
\end{abstract}

\section{INTRODUCTION}

The typical autonomous systems consist of two fundamental core components: perception and control. The perception component engages with the environment through sensors like cameras and LiDAR, capturing RGB imagery and point cloud data. This data is then utilized to classify objects and relevant features, creating a representation of the surrounding environment. Following this, the control component assesses the perceived environment to devise a trajectory for the system and compute the necessary actuation commands, thereby altering the system's state (for clarity in this exposition, planning and decision-making are considered as facets of the control component).

Advances in deep learning have significantly improved perception performance \cite{he2016deep,redmon2018yolov3}. Nonetheless, achieving perfect perception remains challenging due to inherent model limitations, leading to inaccuracies that can mislead the downstream control component into making erroneous decisions. 
While autonomous systems integrating neural network-based perception and control components \cite{pmlr-v205-karkus23a} have shown promising performance, they necessitate differentiability between perception and control components. However, the differentiable control components often have limitations and do not yet match the performance of robust, hand-crafted algorithms employed by traditional control components that proficiently generate actions based on the current state and motion predictions, particularly for complex tasks.

In this vein, formal specifications offer a path to precisely describe safety requirements alongside other intricate properties and have found success in the verification and synthesis of control components within autonomous systems \cite{kress-gazit07,conner07rsj,Kloetzer08,wongpiromsarn10hscc,Tabuada04lineartime,Girard09Hierarchical,karaman09cdc,Castro:2013:CDC,Wongpiromsarn:2021:Minimum}. Numerous tools have also emerged to formally specify and verify simpler machine learning (ML)-based systems \cite{dreossi2019verifai,fremont2020formal,schmitt2022nureality,ivanov2019verisig}, yet their applicability to ML-based perception components remains constrained due to challenges in formally defining properties mirroring human-level perception \cite{dreossi2018semantic}. The literature includes efforts towards formal verification on less complex functions within neural networks; for instance, in~\cite{ivanov2019verisig}, the focus is on verifying or specifying properties of the activation function's output as opposed to the neural network itself, while in~\cite{dreossi2019verifai}, formal specification and verification yield counter-examples for system-level safety rules, albeit the clear linkage between perception models and counter-examples necessitates human analysis. Moreover, as expounded in \cite{dreossi2018semantic,badithela2021leveraging}, not all misclassifications bear equal weight; some possess a higher propensity to trigger system-level failures with graver repercussions. This underscores the imperative for a training framework endowed with system-level specifications and contextual semantics.

This paper introduces a novel approach for embedding system-level safety regulations into the perception components of autonomous systems through reinforcement learning, ensuring that safety objectives are considered during the training process. Our primary contributions are as follows:

\begin{enumerate}
    \item We develop an algorithm that combines task rewards, policy gradients, and system-level safety specifications to create a feedback-learning training framework. This framework embeds system-level safety requirements within the perception component, ensuring compliance with formally articulated safety rules during training.

    \item We present an object detection model training pipeline capable of handling non-differentiable components (e.g., black-box simulators and controllers) in the loop. This pipeline demonstrates the versatility of the proposed algorithm in accommodating non-linearities, discontinuities in safety rules, and realistic constraints within the training environment.

    \item We conduct a rigorous evaluation using the high-fidelity simulator CARLA as a primary testbed to fine-tune state-of-the-art pre-trained perception modules to improve system-level safety. Results highlight the algorithm's efficacy in closely coupling the perception and control modules to enhance system-level safety.
\end{enumerate}

\section{Preliminaries}\label{sec: prelim}

\subsection{Reinforcement Learning Basics}\label{subsec: RL_basics}

The standard method for defining \textit{(deep) reinforcement learning (RL)} is as a decision-making problem where an agent interacts with an environment to achieve a goal \cite{sutton2018reinforcement,mnih2013playing,mnih2015human}. This decision process is typically formulated as a \textit{Markov Decision Process (MDP)} denoted as the tuple \((S, A, P, R, \gamma)\) where $S$ is a set of states, $A$ is set of actions, $P$ is the \textit{state transition probability} where $P(s' | s, a)$  is the probability given state $s$ and taken action $a$ to state $s'$, $R$ is the \textit{reward function} where  $R(s, a, s')$  provides a scalar reward signal in response to the agent's action and $\gamma \in [0,1]$ is a \textit{discount factor}.

The agent interacts with the environment by observing the current state at time \( t \), \( s_{t} \in S \), selecting an action, \( a_{t} \in A \), receiving a reward, \( r_{t} = R(s_{t}, a_{t}, s_{t+1}) \) from the environment, and transitioning to a new state, \( s_{t+1} \) based on $P(s' | s, a)$.  A \textit{trajectory}, denoted \( \tau \), is a sequence of states and actions, \( \tau = [s_0, a_0, s_1, a_1, \ldots, s_T, a_T] \), where \( T \) is the time horizon (possibly infinite). The \textit{cumulative reward} or \textit{return} over a trajectory is often denoted by overloading the reward function notation as \( R(\tau) = \sum_{t=0}^{T} \gamma^{t} r_{t} \) for the discounted return. A policy is a strategy followed by the agent to select actions given a state. Policies can be deterministic, \( a_{t} = \mu_{\theta}(s_{t}) \), or stochastic \( a_{t} \sim \pi_{\theta}(\cdot|s_{t}) \) where \( \theta \) are the parameters of the policy.  In deep RL, \( \theta \) is typically the parameters of a neural network.  The objective in RL is to learn a policy that maximizes the expected cumulative reward, \( J(\pi) = E_{\tau \sim \pi}[R(\tau)] \), giving the optimal policy \( \pi^{*} = \arg \max J(\pi) \).


\textit{Policy gradient} (PG) methods, \cite{williams1992simple, sutton1999policy}, are a class of algorithms in RL that directly optimize the policy in a parameterized form through gradient ascent (descent),
$\theta_{k +1} = \theta_{k} + \alpha \nabla_{\theta} J(\pi_{k}) |_{\theta_{k}}$,
where \( \alpha \) is the learning rate and \( \nabla_{\theta} J(\pi_{k}) |_{\theta_{k}} \) is the gradient of the expected return with respect to the policy parameters evaluated at the current policy parameters. With the log-derivative trick \cite{williams1992simple, sutton1999policy,openai_spinningup_2018}, the gradient can be estimated by 

$$\nabla_{\theta} J(\pi_{\theta})  = \mathbb{E}_{\tau \sim \pi_{\theta}}{\sum_{t=0}^{T} \nabla_{\theta} \log \pi_{\theta}(a_t |s_t) R(\tau)}.$$

\subsection{System-Level Safety Specifications}\label{subsec: rules and rulebook}
We employ the rulebook formalism introduced in \cite{Censi:2019:Liability} to precisely specify system-level safety requirements. At the core of this framework is the concept of defining individual rules, each corresponding to a
distinct aspect of safety (e.g., safety of humans vs. rules of the road), and prioritizing these rules based on their relative importance. For the purpose of this paper, we focus on utilizing the individual rules without considering their prioritization.

A rule is evaluated over a set $\Xi$ of realizations.
In our context, a \textit{realization} is a sequence of world states that encompass the state of the system as well as all pertinent objects and features in the environment.
We denote a realization as \( \mathbf{x} = [x_0, x_1, \ldots, x_T] \), where each \( x_t \) represents the world state at time \( t \).
A rule is a function
\( \mathit{rb}: \Xi \mapsto \mathbb{R}_{\geq 0} \), mapping each realization to a non-negative value representing the severity of rule violation.
Given two realizations \( \mathbf{x_{1}}, \mathbf{x_{2}} \in \Xi \), \( \mathit{rb}(\mathbf{x_{1}}) > \mathit{rb}(\mathbf{x_{2}}) \) implies that \( \mathbf{x_{1}} \) violates the rule more severely than \( \mathbf{x_{2}} \). \( \mathit{rb}(\mathbf{x_{1}}) = 0 \) indicates full compliance with the rule by realization \( \mathbf{x_{1}} \).

\section{Proposed Framework}\label{sec: framework}

\subsection{Components}\label{subsec: framework overview}
\begin{figure*}[ht]
    \centering
    \includegraphics[trim=0cm 0cm 0cm 0cm, clip,width=0.75\textwidth,keepaspectratio]{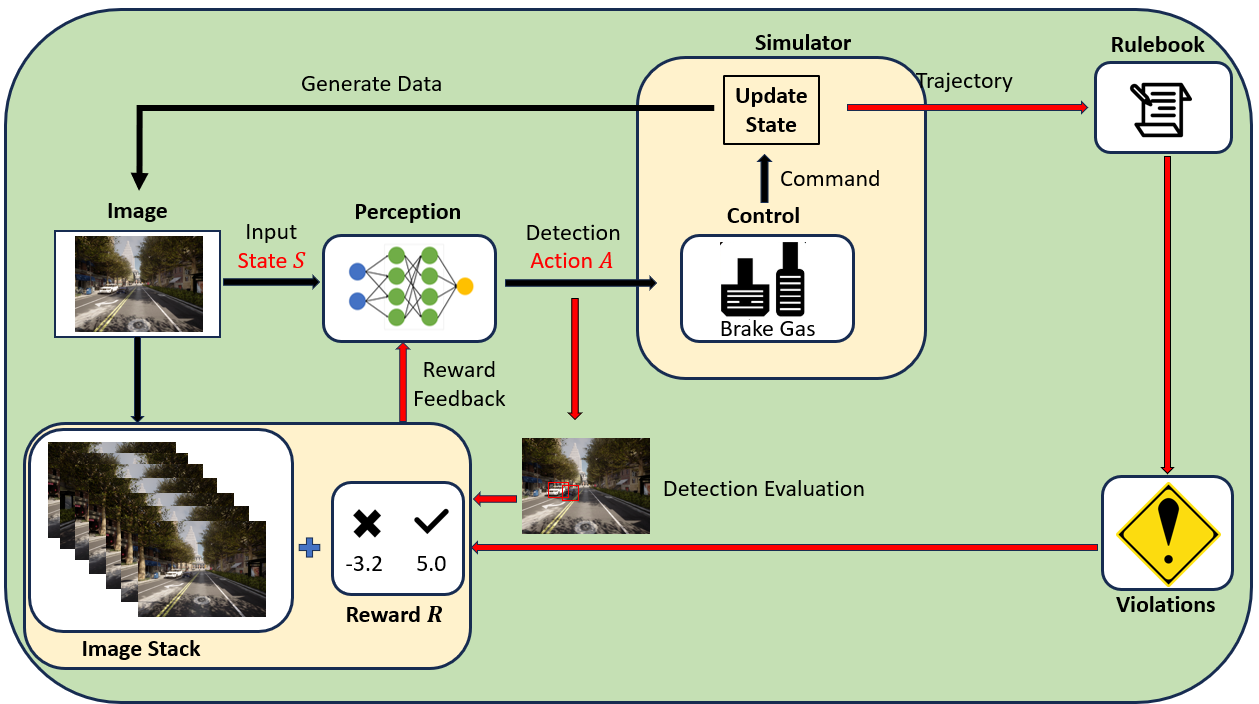}
    \caption{Overview of the proposed feedback learning framework that incorporates both traditional vision-based metrics and safety metrics derived from a rulebook to train or fine-tune an object detection model in the perception component of an autonomous system; the object detection model is reframed as a reinforcement learning agent where, input images are considered as state, bounding box and class outputs are considered as action, and a combination of object detection and safety metrics is considered as the reward function.}
    \label{fig: framework}
\end{figure*}

This section outlines our proposed framework for training the perception component to be aware of system-level objectives through feedback learning. The framework consists of (i) perception, (ii) control, (iii) simulator, and (iv) rulebook, as shown in Fig. \ref{fig: framework}. These components are organized in order to provide a simulation pipeline for collecting trajectories as the ``inner loop'' of a training cycle. 


\subsubsection{Perception Component}\label{subsubsec: perception}
We focus on probabilistic machine learning models for the perception component, defined as a function $PC_{\theta}$, where $\theta$ represents the tunable model parameters.
Let \( \mathbf{S}_{pc} \) denote the set of all possible input sensor data (e.g., the set of all possible RGB images matrices) and \( \mathbf{E} \) denote the set of all possible perceived environments (e.g., the set of all possible configurations of 3D bounding boxes and their corresponding object types).
We define $\mathbf{P} = [0,1]^n$ as the set of probability vectors, where $n \in \mathbb{N}$ corresponds to the dimensionality of $\mathbf{E}$ (e.g., the maximum number of output bounding boxes). 

Given the model parameters $\theta$, the function $PC_{\theta}$ takes sensor data $\mathbf{s}_{pc} \in \mathbf{S}_{pc}$ and an environment $\mathbf{e} = [e_1,e_2,\ldots,e_n] \in \mathbf{E}$ as inputs.
It returns a probability vector $\mathbf{p} = [p_1,p_2,\ldots,p_n] \in \mathbf{P}$, where each $p_i$, $i \in \{1, \ldots, n\}$, represents the probability of the corresponding element $e_i$.
To emphasize that these probabilities are conditioned on the sensor data, we denote this function as $PC_{\theta}(\mathbf{e}|\mathbf{s}_{pc}) \in \mathbf{P}$, rather than using the usual function notation $PC_{\theta}(\mathbf{s}_{pc}, \mathbf{e})$.
For instance, in an object detection model, $PC_{\theta}(\mathbf{e}|\mathbf{s}_{pc})$ represents the probabilities of detecting a specific set of objects $\mathbf{e}$ given the input image $\mathbf{s}_{pc}$.

Note that unlike deterministic models that directly predict a specific environment $\mathbf{e} \in \mathbf{E}$, probabilistic models output a vector of probabilities, reflecting the uncertainty of each possible perceived environment $\mathbf{e} \in \mathbf{E}$ based on the input $\mathbf{s}_{pc}$ and the model parameters $\theta$.
This type of model is widely used in computer vision tasks, such as object detection \cite{chen2021pix2seq} and image segmentation \cite{jain2023oneformer} models based on Vision Transformers (ViT) \cite{dosovitskiy2020image}.


\subsubsection{Control Component}\label{subsubsec: ctrl}

The primary function of the control component is to compute actuation commands based on the perceived environment from the perception component, as delineated in Figure~\ref{fig: framework}. Formally, this component is defined as a function \( CTRL: \mathbf{E} \times \mathbf{P} \times \mathbf{S}_{sys} \mapsto U_{\text{ctrl}} \),  where \( \mathbf{S}_{sys} \) is the state space of the autonomous system, \( U_{\text{ctrl}} \) denotes the set of all possible actuation commands. As defined earlier in the perception component, $\mathbf{P} = [0,1]^n$ represents the uncertainty space over the perceived environment $\mathbf{E}$, with the same dimensionality $n$ as $\mathbf{E}$. This formulation allows the control component to translate the perceived environment, along with its associated uncertainty and current system state, into executable actuation commands. One possible way to define the perceived environment $\mathbf{e} = [e_1, \ldots, e_n]$ is by selecting one that maximizes the overall likelihood of the observed environment. For instance, $\mathbf{e}$ can be chosen as the set of object detection results that maximize the product of their individual probabilities: $\mathbf{e} = \arg \max_{\mathbf{e} \in \mathbf{E} } {\prod_{i=1}^{n}PC_{\theta}(e_i|\mathbf{s}_{pc})}$. Here, $PC_{\theta}(e_i|\mathbf{s}_{pc})$ is used, with some abuse of notation, to denote the specific probability $p_i$ within the probability vector $PC_{\theta}(\mathbf{e}|\mathbf{s}_{pc}) = [p_1, \ldots, p_n]$. Essentially, $PC_{\theta}(e_i|\mathbf{s}_{pc})$ refers directly to the probability assigned to the element $e_i$ of the perceived environment $\mathbf{e}$, given the input sensor data $\mathbf{s}_{pc}$.
It is important to note that this is just one possible approach for defining the perceived environment $\mathbf{e}$; other methods may be suitable depending on the application or specific requirements.
The system state $\mathbf{s}^{sys} \in \mathbf{S}_{sys}$ includes information such as position, velocity, and acceleration. Based on these inputs and the control logic, the control component then outputs the actuation command $u \in U_{\text{ctrl}}$. 

\subsubsection{Simulator}\label{subsubsec: simulator}
The simulator processes the actuation commands from the control component and updates the simulated world. Over multiple consecutive simulation steps $t \in \mathbb{N}$, it generates a sequence of input sensor data $[\mathbf{s}_1,...,\mathbf{s}_t]$, where $\mathbf{s}_i \in \mathbf{S}_{pc}$, as well as the corresponding sequence of world states $[x_1,...,x_t]$, which forms a realization for rulebook. Each world state $x_i = (\mathbf{s}_{sys, i}, \mathbf{e}_{gt, i})$ for $i \in \{1, \ldots, t\}$ includes the system state $\mathbf{s}_{sys,i} \in \mathbf{S}_{sys}$ and the ground truth of observed environment $\mathbf{e}_{gt,i} \in \mathbf{E}$.

\subsubsection{Rulebook Component}\label{subsubsec: rulebook}

We utilize the concept of rules as defined in Section \ref{sec: prelim} to specify system-level safety objectives. The values returned by the rules are interpreted as the violation scores for a realization generated by the simulator, where higher scores correspond to more severe violations. The rulebook component containing $K$ rules takes the realization $\mathbf{x}$ from the simulator as input and produces \( \mathit{rb}(\mathbf{x}) = [v_{1},v_{2},...,v_{K}] \in \mathbb{R}^{K}_{\geq 0} \), where each \( v_{i} \) represents the violation score of the $i$th rule.


\begin{remark}
To ensure that the controller behaves reasonably, we assume that when the controller is provided with the ground truth of the environment $\mathbf{e} \in \mathbf{E}$, the control component results in zero violations across all rules. This assumption implies that as the perception component improves, leading to more accurate environmental understanding, the overall system's safety also improves accordingly.
\end{remark}

\subsection{Reinforcement Learning Formulation}
Using the framework outlined in Section \ref{subsec: framework overview}, we aim to train the probabilistic machine learning model with system-level safety requirements.
We formulate this framework as a Reinforcement Learning process modeled as an MDP \((S, A, P, R, \gamma)\), where
$S = \mathbf{S}_{pc}$ represents the state space corresponding to the input of the perception component,
$A = \mathbf{E}$ represents the action space corresponding to the set of all possible environments,
$P$ represents the state transition probabilities, which depend on the state, action, and control logic. 
Unlike traditional RL frameworks where rewards are directly generated from a reward function $R$, our approach derives rewards from the violation scores produced by the rulebook component. As in typical RL reward design, there are various ways to derive these rewards. Generally, the reward should be monotonic with respect to the violation scores, though there is no specific requirement on its exact form. An example of reward design is provided in Section \ref{subsec:Reward Design}.
Finally, we consider a stochastic policy defined as $\pi_{\theta}(\cdot|s_{t}) = PC_{\theta}(\mathbf{e}|\mathbf{s}_{pc})$,
where the policy outputs a probability distribution over possible environments $\mathbf{e}$, given the sensor data $\mathbf{s}_{pc}$.








\subsection{Policy Gradient Algorithm}
Using the framework and reinforcement learning formulation described earlier, we apply the Policy Gradient Algorithm \cite{pinto2023tuning} to train a probabilistic model $\pi_{\theta}(\cdot|s_{t})$, which is parameterized by $\theta$. This training process, outlined in Algorithm \ref{alg:reward-optimization}, takes the model parameter $\theta$ and learning rate $\alpha$ as inputs and trains the model over a specified number of epochs (max\_epoch) using gradient $G_r$. In each epoch, it samples max\_traj trajectories via Algorithm \ref{alg:sample-traj}, based on
a given model parameter $\theta$ and initial state $s_0$. Steps 1-4 are the corresponding components we define in \ref{subsec: framework overview}. 

\setlength{\textfloatsep}{1pt}
\begin{algorithm}
\begin{algorithmic}
    \caption{Training with Rewards over Trajectories}
        \Function{TrainingWithReward}{$ \theta, \alpha $}
        \For{epoch = 1 to max\_epoch}
        \State{$G_r = 0$}
        \For{trajectory = 1 to max\_traj}
        \State{Randomly choose a initial state $s_0$ from $S$}
        \State \( \mathcal{T}, \mathcal{A},
        R\gets \text{SampleTrajectory}(\theta, s_{0}) \)
        \State{Loss = $\frac{1}{N} \sum_{i=1}^{T}   \sum_{j = 1}^{n}\log \pi( A_{i,j}|\mathcal{T}_i,\theta)R_{i,j}$}
        \State{\( G_r += \nabla_{\theta} \text{Loss}\) }
        \EndFor
        \State{Update model parameter \( \theta = \theta + \alpha G_r \)}
        \EndFor  
        \State \Return $\theta$
        \EndFunction
    \label{alg:reward-optimization}
\end{algorithmic}
\end{algorithm}

\begin{algorithm}
    \begin{algorithmic}
    \caption{Sample Trajectory for Training}
            \Function{SampleTrajectory}{$\theta, s_{0}$ }  
        \State{Initalize $\mathcal{T}, \mathcal{A}, R, V$ as empty set}
        \For{$t = 0$ to $T$}
            \State{$\#$ 1. Perception Component step}
            \State{Get current $s_{t}$ from simulator}
            \State{Select action $a$ based on $\mathbf{p} = PC_{\theta}(a|s_{t})$}
            \State{$\#$ 2. Control Component step}
            \State{Get control signal $u_t = CTRL(a,\mathbf{p},s_{sys})$}
            \State{$\#$ 3. Simulation step}
            \State{Get the current world state $x_t$ from simulator}
            \State{$\mathbf{x}$.append($x_t$) then run one step simulation with $u_t$}
            \State{$\#$ 4. Rulebook Component step}
            \State{Compute violation $\mathbf{v} = \mathit{rb}(\mathbf{x})$, $V$.append($\mathbf{v} $)}


            \State{$\mathcal{T}$.append($s_t$), $\mathcal{A}$.append($a$)}
        \EndFor
        \State{$\#$ 5. Reward Calculation}
        \State{Calculate reward $R$ from $V$}
        \State \Return \( \mathcal{T}, \mathcal{A}, R \)
        \EndFunction
    \label{alg:sample-traj}
    \end{algorithmic}
\end{algorithm}


\section{Implementation Details}\label{sec:implementation detail}

\subsection{Perception Component}\label{subsec:perception implementation}
To demonstrate our approach, we employ the state-of-the-art Maximum Likelihood Estimation (MLE)-based object detection model, PIX2SEQ \cite{chen2021pix2seq}. PIX2SEQ formulates object detection as a language modeling task, which encodes bounding box coordinates and class labels as vocabulary tokens and predicts these tokens in sequence. 
The output sequence includes tokenized values \( [x_{\min}, x_{\max}, y_{\min}, y_{\max}, \mathit{cl}] \), which represent the bounding box coordinates \(x_{\min}, x_{\max}, y_{\min}, y_{\max}\) and class label \(\mathit{cl}\).
Although PIX2SEQ serves as a specific implementation for this demonstration, our framework is flexible and can be applied to other probabilistic object detection models as well.


Typically, tokens are generated by selecting the one with the largest likelihood (arg max sampling) from the softmax layer's output. In our experiment, we treat the softmax output as a probability distribution over actions, denoted as $\pi_{\theta}(\cdot|\mathbf{s}_{pc})$, and choose the token randomly according to the probability of each word in the vocabulary. The RL training process then updates the policy network based on system-level safety requirements.

In our implementation, input images are resized to \( 384 \times 384 \) pixels, with 80 possible detection classes. Each token corresponds to an element from the vocabulary $\text{Voc} = \{1,\ldots,...,384, cl_1, \ldots, cl_{80}\}$, where $cl_i$ is the $i$th detection class. The model is configured with ViT-B/16 as the image encoder (pretrained on IMAGENET-1K) and 6-layer auto-regressive Transformer decoder. Bounding box coordinates are discretized into 384 buckets, with an additional token for the class label.  The model is fine-tuned on the COCO dataset \cite{lin2015microsoft} to serve as our pretrained detection model for the perception component.

Finally, the perception component integrates depth data with each image by fusing PIX2SEQ's tokenized output \( [x_{\min}, x_{\max}, y_{\min}, y_{\max}, \mathit{cl}] \) with depth information to produce a set $\mathbf{e}$ of detected 3D objects. Each object $e = (x_c,y_c,z_c,w, h, d,v) \in \mathbf{e}$ is defined by its center $(x_c,y_c,z_c)$, dimensions (width $w$, height $h$, depth $d$), and speed $v$. The perception component outputs the set of detected objects $\mathbf{e}$ and the associated probability vector $\mathbf{p}$.

\subsection{Simulator}
We use CARLA \cite{Dosovitskiy17} as our simulation platform to conduct realistic  simulations and address a wide range of corner cases. 
We run the simulation in synchronous mode with a fixed time-step of 0.1 seconds. The ego vehicle is spawned at a random location within the simulator, and its environment is monitored using sensors mounted on the vehicle, including an RGB camera, a depth camera, and radar. The controller utilizes the detection results from these sensors to generate actuation commands. Additionally, we spawn additional objects to populate the simulator, including 35 different types of vehicles and 50 different types of pedestrians. Our ego vehicle operates using the previously described components, while the other objects function under the CARLA's built-in auto-pilot mode. 

\subsection{Rulebook Component}\label{subsec: rulebook implementation}
We consider a rulebook with four rules:
$r_1$ (collision avoidance),
$r_2$ (obstacle clearance),
$r_3$ (absence of unnecessary brake) and $r_4 $ (progress maintenance). 
These rules are assumed to be additive, meaning they evaluate violations at each state in a realization. The total violation for each rule is the sum of violations across all states in the realization. 
Each rule $r_k$ is represented by a function $\mathit{rb}_k:  X \to \mathbb{R}_{\geq 0}$, where
$X = \mathbf{S}_{sys} \times \mathbf{E}$ is the set of world states.
A world state $x = (\mathbf{s}_{sys}, \mathbf{e})$ consists of the system state
$\mathbf{s}_{sys} = (x_e, y_e, \theta_e, v_e, a_e)$, with
$(x_e, y_e, \theta_e)$ as the ego vehicle's pose and $v_e$ and $a_e$ denoting its speed and acceleration,
and the set $\mathbf{e} \in \mathbf{E}$ of ground truth 3D bounding boxes.
The function $\mathit{rb}_k(x)$ then returns the violation score for the system state $\mathbf{s}_{sys}$ and the corresponding ground truth environment $\mathbf{e}$ according to rule $r_k$.

Let $O \subseteq \mathbf{e}$ denote the set of objects in the same lane and in front of the ego vehicle.
For each $o_i \in O$, let $d_i$ denote the distance between $o_i$ and the ego vehicle, let $v_i$ denote the signed speed of $o_i$.
$v_i \geq 0$ indicates that $o_i$ is moving in the same direction as the ego vehicle, while $v_i < 0$ means that $o_i$ is moving in the opposite direction.

First, consider the collision avoidance rule.
We define the violation metric for this rule based on the kinetic energy of the ego vehicle at collision, which is a typical measure of injury severity.
Specifically, for each $o_i \in O$, 
$\mathit{rb}_{1,i}(x) = v_e^2$ if $d_i < \epsilon$ (for some small $\epsilon \geq 0$); otherwise $\mathit{rb}_{1,i}(x) = 0$
We then define the collision avoidance rule as $\mathit{rb}_{1}(x) = \sum_{o_i \in O} \mathit{rb}_{1,i}(x)$.

To define the obstacle clearance rule, we utilize the Responsibility-Sensitive Safety (RSS) condition \cite{hasuo2022responsibility}, which is a conservative condition that guarantees collision avoidance. Assume that $v_i \geq 0$ for all $o_i \in O$.
Additionally, to simplify the presentation, we assume that the maximum response time that the ego vehicle might take to initiate the required braking is 0.
Then, for each $o_i \in O$, we define the required clearance

\begin{equation*}
c_i = \left\{ \begin{array}{ll} 
\max\left( 0, \frac{v_e^2}{2 a_{brake}} - \frac{v_i^2}{2a_{brake,i}} \right) &\text{ if } o_i \text{ is a vehicle}\\
\frac{v_e^2}{2 a_{brake}} &\text{ otherwise}
\end{array}\right.
\end{equation*}
where $a_{brake} > 0$ is the comfortable braking rate for the ego vehicle, and
$a_{brake,i} \geq a_{brake}$ corresponds to the maximum braking rate for object $o_i$.
The clearance rule is then defined as the sum of the differences between the actual and the required clearance:
\begin{equation*}
\mathit{rb}_2(x) = \left\{\begin{array}{ll}
\sum_{o_i \in O} \max(0, c_i - d_i) &\text{ if } O \not= \emptyset\\
0 &\text{ otherwise}
\end{array}\right.
\end{equation*}

We will also utilize the RSS safety condition in defining the absence of unnecessary brakes.
Let $\tau > 0$ denote a constant representing a time buffer. The distance the ego vehicle is expected to travel during this time is then $v_e \tau$.
We then define the absence of unnecessary brake rule as

\begin{equation*}
\mathit{rb}_3(x) = \left\{ \begin{array}{ll}
\max(0, -a_e) &\text{ if } d_i > c_i + v_e \tau + \frac{1}{2}a_{brake} \tau^2\\
 & , \forall o_i \in O\\
0 &\text{ otherwise}
\end{array}
\right.
\end{equation*}

Finally, we define the progress maintenance rule to ensure that the ego vehicle keeps moving if the environment is sufficiently clear. Since the progress in one simulation time step $\Delta t$ depends on acceleration,  we assess progress by comparing the ratio of actual acceleration to the target acceleration.
\begin{equation*}
    \mathit{rb}_4(x) = \left\{ \begin{array}{llll}
     \max(r - \frac{a_e}{a_{target}} ,0 )&\text{ if } O = \emptyset \text{ and } a_{target} > 0\\
    \max(r - \frac{a_e}{a_{target}} ,0) &\text{ if }  d_i > c_i + v_e \tau + \frac{1}{2}a_{brake} \tau^2\\
         & , \forall o_i \in O \text{ and } a_{target} > 0\\
    0 &\text{ otherwise}
\end{array}    
\right.
\end{equation*}
where $ 0< r \leq 1$ is a predefined ratio which measures the allowed ratio between $a_{target}$ and $a_e$, where $a_{target}$ is decided based on equation 
\ref{equ:a target}, $a_{max}$, $a_{min}$ are the maximum acceleration rate and maximum braking rate of the ego vehicle, respectively. 

\begin{equation*}\label{equ:a target}
    a_{target} = \left\{ \begin{array}{lll}
     \min(a_{max},\frac{v_{lim}-v_e}{\Delta t})&\text{ if } O = \emptyset\\
    \min(a_{max},\frac{v_{max} - a_{brake} \Delta t - v_e}{\Delta t}) & \text{ if } \forall o_i \in O\\
     \text{\qquad } d_i > c_i + v_e \tau + \frac{1}{2}a_{brake} \tau^2 &\\
    0 &\text{ otherwise}
\end{array} 
\right.
\end{equation*}
Here, $v_{lim}$ denotes the speed limit.
Finally, we define the maximum permissible speed $v_{max} = \min_{o_i \in O} v_{max,i}$ if $O \not= \emptyset$ and $v_{max} = \infty$ otherwise. Here, $v_{max,i}$ is the maximum speed for the ego vehicle to not violate $r_2$ with respect to object $o_i \in O$ and is defined as
\begin{equation*}
\label{equ:1}
v_{max, i} = \left\{\begin{array}{ll}
\sqrt{2a_{brake} \left(d_i + \frac{v_i^2}{2a_{brake,i}}\right)} &\text{ if } o_i \text{ is a vehicle}\\
\sqrt{2a_{brake}d_i} &\text{ otherwise}
\end{array}
\right.
\end{equation*}

Given a realization $\mathbf{x} = [x_0,...,x_T]$, its violation score according to rule $r_i$ is the sum over the violation scores across all states in the realization: $\mathit{rb}_i(\mathbf{x}) = \sum_{t \in \{1, \ldots, T\}} \mathit{rb}_i(x_t)$.

\subsection{Control Component}
We consider a longitudinal controller $CTRL(\mathbf{e},\mathbf{p},\mathbf{s}_{sys})$, which takes as input $\mathbf{e}$ and $\mathbf{p}$---the 3D bounding boxes with speed information and corresponding probabilities from the perception component as described in Section \ref{subsec:perception implementation}---as well as the system state $\mathbf{s}_{sys} = (x_e, y_e, \theta_e, v_e, a_e)$, defined as in section \ref{subsec: rulebook implementation}.  
The controller then produces an acceleration command $a \in [-a_{min}, a_{max}]$, where $a_{min}$ is the maximum braking rate allowed by the simulation, and $a_{max}$ is the maximum acceleration of the ego vehicle.

From the set $\mathbf{e}$, we can compute the set $O \subseteq \mathbf{e}$ and define the maximum permissible speed $v_{max}$ as in Section \ref{subsec: rulebook implementation}.
Suppose the controller updates the acceleration command every $\Delta t$ seconds (i.e., each simulation time step) and the comfortable braking $a_{brake}$ defined in Section \ref{subsec: rulebook implementation} satisfies $a_{min} \geq a_{brake} > 0$.
Then, in the next simulation time step, the minimum speed among $v_{max,i}$ after applying full braking with $a_{brake}$ is given by $v_{max} - a_{brake}\Delta t$.
Thus, we define the controller as

\begin{equation*}\label{equ:control}
\mathbf{CTRL}(\mathbf{e},\mathbf{p},\mathbf{s}_{sys}) = \left\{
\begin{array}{ll}
\min\{a_{max}, \frac{v_{lim} - v_e}{\Delta t}\} &\\
\text{ \quad if } O = \emptyset &\\
-a_{min} & \\
\text{ \quad if } v_e > v_{lim} & \\
\min\{a_{max}, \frac{v_{max} - a_{brake}\Delta t - v_e}{\Delta t} \} &\\
\text{ \quad otherwise} &
\end{array}
\right.
\end{equation*}
We now show that the proposed controller ensures that the rules $r_1$, $r_2$, $r_3$ and $r_4$ with violation scores 0.

\begin{lemma}
\label{lem:r1r2}
Suppose (1) $\mathit{rb}_1(x_t) = \mathit{rb}_2(x_t) = 0$ at arbitrary time $t$, (2) the set $O$ of objects in the same lane and in front of the ego vehicle remains the same at time $t$ and $t + \Delta t$, and (3) $a_{brake,i}$ is the maximum braking rate of object $o_i$.
Then, applying $CTRL$ at time $t$ ensures that $\mathit{rb}_1(x_{t+\Delta t}) = \mathit{rb}_2(x_{t+\Delta t}) = 0$ at time $t+\Delta t$.
\end{lemma}

\begin{lemma}
\label{lem:r3}
Suppose at an arbitrary time $t$, all the assumptions of Lemma \ref{lem:r1r2} are satisfied and additionally assume that (4) $v_e \leq v_{lim}$ at time $t$, and (5) $\tau \geq \Delta t$. Then, $\mathit{rb}_3(x_t) = 0$.
\end{lemma}

\begin{lemma}
\label{r4}
Suppose at an arbitrary time $t$, all the assumptions of Lemma \ref{lem:r1r2} and \ref{lem:r3} are satisfied, then $\mathit{rb}_4(x_t) = 0$ at arbitrary time $t$.

\end{lemma}
\begin{corollary}
\label{lem:r1r2r3}
Suppose at time 0, $\mathit{rb}_1(x_0) = \mathit{rb}_2(x_0) = \mathit{rb}_3(x_0) = \mathit{rb}_4(x_0) = 0$ and all the assumptions of Lemma \ref{lem:r1r2}, \ref{lem:r3} and $\ref{r4}$ are satisfied. Then $\mathit{rb}_1(\mathbf{x}) = \mathit{rb}_2(\mathbf{x}) = \mathit{rb}_3(\mathbf{x}) = \mathit{rb}_4(\mathbf{x}) = 0$
\end{corollary}

\subsection{Reward Design}\label{subsec:Reward Design}
We have two parts of the reward: the perception reward $r_{pc}$ and rulebook $r_{rb}$ reward. So, the total reward for each token would be
$$r = \beta r_{pc} + (1-\beta) r_{rb}$$ 
where $\beta \in [0,1]$ is a tunable parameter to balance between perception reward and rulebook reward. 

Inspired by \cite{pinto2023tuning}, we use Intersection over Union (IoU) as a metric to build perception reward. For each bounding box, if the IoU with ground truth exceeds a given threshold and the label is correct, we assign a perception reward of $r_{pc} = w_{percp}$ for all 5 tokens, where $w_{percp}$ is a weight parameter. Otherwise, we replace the token with ground truth and assign the same reward during training.

On the other hand, the rulebook reward is calculated from the violation scores. 
Given a bounding box and its associated realization $\mathbf{x}$, if it is a wrong detection result and triggers the violation with rule $r_i$, then $r_{rb_i} = \mathit{rb}_i(\mathbf{x})$; otherwise $r_{rb_i} = 0$. Here the realization at time step $t$ is $\mathbf{x} = [x_t,x_{t+1},x_{t+2},...,x_{T}]$, thus reward is designed to measure the influence of simulation from now to the end with rules. The total rulebook reward is $r_{rb} = \sum_{i=1}^{K}r_{rb_i}(\mathbf{x})$. For a bounding box with 5 tokens, if $r_{rb} = 0$, then the original detection result is kept; otherwise, we replace it with the ground truth during training.



\section{Experimental Results And Discussion}

This section details the training methodology, analyzes experimental results, and evaluates the proposed framework and implementations from section \ref{sec: framework} and \ref{sec:implementation detail}.

\subsection{Training Methodology}






The ViT based perception component PIX2SEQ was first pre-training on the COCO dataset \cite{ lin2014microsoft}. 
The pre-training was executed over \( 100 \) epochs, with fixed learning rate \( \alpha = 0.0001 \). The Adam optimizer was used with a Cross Entropy loss and gradient clipping at \( 1.0 \). The batch size was set to \( 32 \), and the model was trained on a single NVIDIA GeForce RTX 3090 GPU.  Pre-training was performed until the cross entropy loss converged below 0.05.

After getting the pre-trained model, the RL training was conducted using Algorithm \ref{alg:reward-optimization} on a single NVIDIA GeForce RTX 3090 GPU. The training was executed over \( 20 \) epochs, with a learning rate initialized to \( \alpha = 0.0001 \). The Adam optimizer was used with a gradient clipping set to \( 1.0 \). During each epoch, 5 trajectory rollouts were collected, each consisting of 100 simulation steps.  Image mini-batches were extracted from the trajectory rollouts and used to update the model through gradient accumulation across all 5 rollouts.

\subsection{Training Results}
\begin{figure}
    \centering
    \begin{subfigure}{0.3\textwidth}
        \centering
        \includegraphics[width=\textwidth]{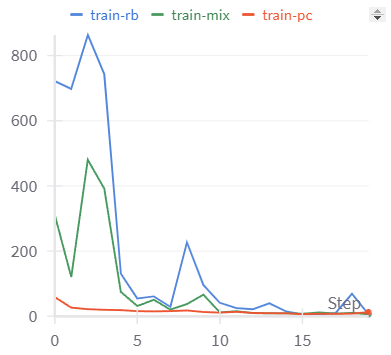} 
        \caption{Total loss}
        \label{fig:subfig1}
    \end{subfigure}
    \hfill
    \begin{subfigure}{0.3\textwidth}
        \centering
        \includegraphics[width=\textwidth]{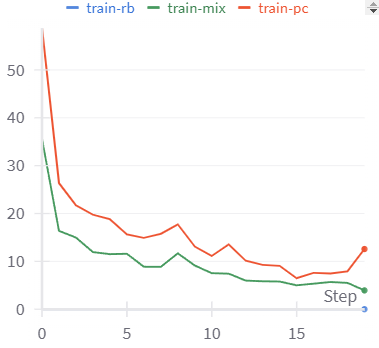} 
        \caption{Perception loss}
        \label{fig:subfig2}
    \end{subfigure}
    \hfill
    \begin{subfigure}{0.3\textwidth}
        \centering
        \includegraphics[width=\textwidth]{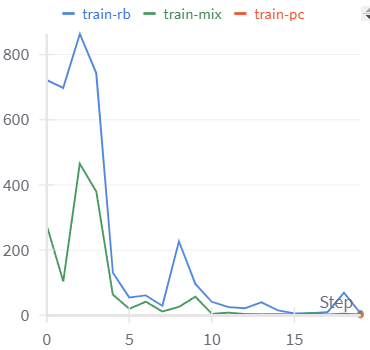} 
        \caption{Rulebook loss}
        \label{fig:subfig3}
    \end{subfigure}
    
    \caption{Training Losses of different settings}
    \label{fig:losses}
\end{figure}

To understand the effect of perception and rulebook rewards, we conducted 3 experiments: training with perception reward only (pc), rulebook reward only (rb) and a combination of both (mix). Figure \ref{fig:losses} shows the training losses of these 3 setups.
Initially, the total loss oscillates as the algorithm explores different policies. This oscillation is primarily driven by the rulebook loss (see Algorithm \ref{alg:reward-optimization}), since the model has not yet been fine-tuned using rulebook reward. In contrast, the perception loss consistently decreases, as the model was fine-tuned on the COCO dataset, allowing detection results to gradually align with the ground truth based on perception reward. After 10 epochs, the model converges to near-zero loss, indicating successful training under the predefined loss functions.

\begin{figure}
        \centering
    \begin{subfigure}{0.35\textwidth}
        \centering
        \includegraphics[width=\textwidth]{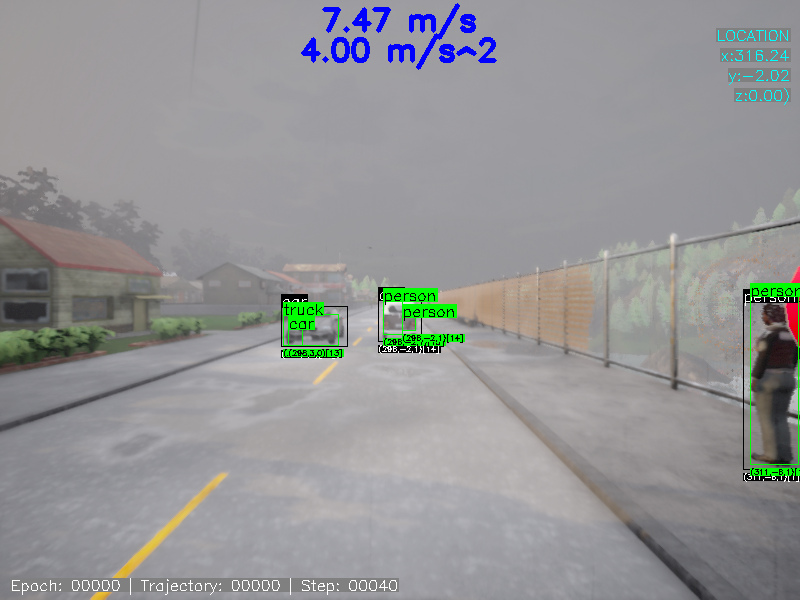} 
        \caption{Detection result at epoch 0}
        \label{subfig:epoch 0}
    \end{subfigure}
    \hfill
    \begin{subfigure}{0.35\textwidth}
        \centering
        \includegraphics[width=\textwidth]{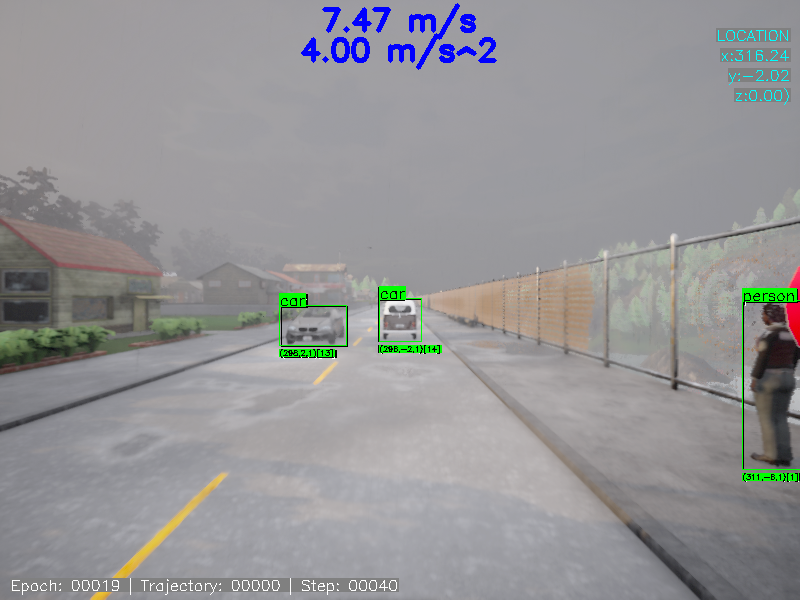} 
        \caption{Detection result at epoch 19}
        \label{subfig:epoch 19}
    \end{subfigure}
    \caption{Detection result comparison among pre-trained and fine-tuned model}
    \label{fig:detection comparision}
\end{figure}

Figure \ref{fig:detection comparision} shows the detection results under the same scenario at training epoch 0 (pre-trained model) and epoch 19 (fully fine-tuned model). Black bounding boxes represent the ground truth, while green bounding boxes show the detections from the perception component. It demonstrates that the training process improves the quality of detection.

\subsection{Evaluation}
\begin{figure}
    \centering
    \begin{subfigure}{0.35\textwidth}
        \centering
        \includegraphics[width=\textwidth]{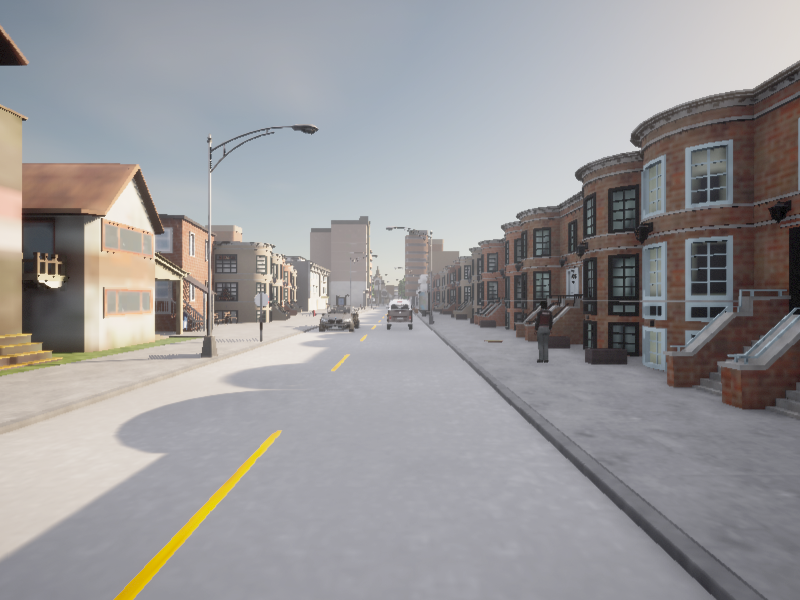} 
        \caption{Simulation under normal weather}
        \label{subfig:normal}
    \end{subfigure}
    \hfill
    \begin{subfigure}{0.35\textwidth}
        \centering
        \includegraphics[width=\textwidth]{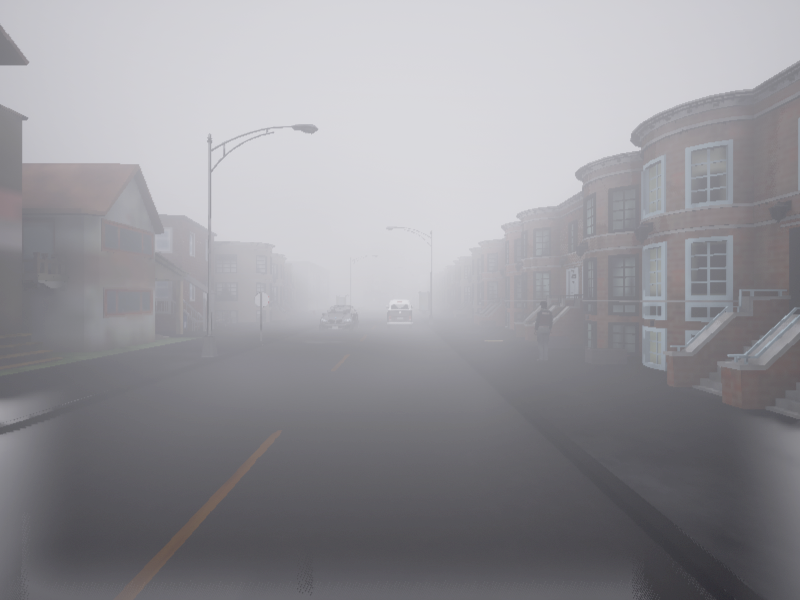} 
        \caption{Simulation under severe weather}
        \label{subfig:severe}
    \end{subfigure}
    \caption{Different weather condition simulation of same scenario}
    \label{fig:weather}
\end{figure}

To evaluate the trained model,  we generate new scenarios using a different map in CARLA than the one used during training. To introduce variability, we spawn the objects with randomly selected types. 
To check the stability of the models, we also evaluate the performance under two weather conditions: normal (fog density = 0) and severe (fog density = 40). 
The model was trained only under normal conditions. Figure \ref{fig:weather} shows an example of these weather conditions.

Table \ref{tab:eva_normal} and \ref{tab:eva_severe} show the results for different weather conditions across the 4 rules.
The tables compare models trained with perception loss only (pc), rulebook loss only (rb), and a combination of both (mix). Table \ref{tab:eva_normal} shows that under normal weather, the rb-only training strategy performs slightly better than both the pc-only and mixed strategies.
Table \ref{tab:eva_severe} indicates that under severe weather, both rb-only and mixed training strategies maintain effective object recognition low violation scores comparable to those achieved under normal weather. This demonstrates the effectiveness of incorporating system-level safety objectives through the rulebook. 

Table \ref{tab:P vs NP} shows the accuracy of prioritized and non-prioritized objects under various weather conditions (fog density = [0, 20, 40, 60]). Here, prioritized objects are those involved in violation score calculations. In our experiments, the rulebook focuses on objects within the same lane and traveling in the same direction. The table indicates that training with rulebook loss (rb and mix) aligns the detection model more closely with system-level safety objectives, improving its focus on prioritized objects. Compared with training with perception loss alone, incorporating rulebook loss helps the perception component focus more on prioritized objects based on safety requirements.

\begin{table}[H]
\caption{Evaluation results under normal weather}
\begin{center}
\label{tab:eva_normal} 
\begin{tabular}{ |c||c|c|c|c|c| } 
 \hline
 & \textbf{$r_1$} & \textbf{$r_2$} & \textbf{$r_3$} & \textbf{$r_4$} & \textbf{total}  \\ 
 \hline \hline
 \textbf{pc} & 42.786 & 56.842& 1.98& 4.006& 105.61
  \\
 \hline
 \textbf{rb} & 25.10 & 51.36 & 0 & 0 & 76.46
 \\
 \hline
 \textbf{mix} & 45.65
 & 54.15 & 0 & 2.48
 & 102.28

  \\
 \hline 
\end{tabular}
\end{center}
\end{table}

\begin{table}[H]
\caption{Evaluation results under severe weather}
\begin{center}
\label{tab:eva_severe} 
\begin{tabular}{ |c||c|c|c|c|c| } 
\hline
 & \textbf{$r_1$} & \textbf{$r_2$} & \textbf{$r_3$} & \textbf{$r_4$} & \textbf{total}  \\ 
 \hline \hline
 \textbf{pc} & 303.59& 171.37& 2.4& 1.94& 479.31
  \\
 \hline
 \textbf{rb} & 47.4 & 81.35 & 0& 0.62& 129.37
 \\
 \hline
 \textbf{mix} & 86.10 & 73.07 & 0.6 & 4.26 & 164.04
  \\
 \hline 
\end{tabular}
\end{center}
\end{table}

\begin{table}[ht]
\caption{Accuracy of prioritized versus non-prioritized objects}
\begin{center}
\label{tab:P vs NP}
\begin{tabular}{|l||c|c|c|c|c|c|}
\hline
 & \multicolumn{3}{c|}{\textbf{Prioritized Acc}} & \multicolumn{3}{c|}{\textbf{Non-Prioritized Acc}} \\ \hline \hline
 & \textbf{rb} & \textbf{pc} & \textbf{mix} & \textbf{rb} & \textbf{pc} & \textbf{mix} \\ \hline
\textbf{weather-0} & 0.84 & 0.54 & 0.78 & 0.58 & 0.66 & 0.68 \\ \hline
\textbf{weather-20} & 0.71 & 0.38 & 0.73 & 0.49 & 0.39 & 0.73 \\ \hline
\textbf{weather-40} & 0.63 & 0.25 & 0.69 & 0.40 & 0.05 & 0.60 \\ \hline
\textbf{weather-60} & 0.28 & 0.13 & 0.33 & 0.06 & 0.03 & 0.03 \\ \hline
\end{tabular}
\end{center}

\end{table}
%








\section{Conclusion}

This work presented a training paradigm that incorporates system-level safety within the perception component of an autonomous system. Utilizing reinforcement learning and a rulebook for training guidance, the framework aims to enhance the alignment of the perception component with safety standards. The algorithm developed combines task rewards with policy gradients, informed by rulebook adherence.
An adaptable training and simulation pipeline was developed to include non-differentiable components, allowing the training to reflect realistic constraints and non-linearities. The CARLA simulator served as the testbed for fine-tuning pre-trained perception component, which showed improved safety metrics compliance.
The evaluations indicated the algorithm's capability to refine detection or classification models within a simulated environment, suggesting its utility in improving autonomous system safety.







\bibliographystyle{IEEEtran}
\bibliography{references}

\end{document}